\documentclass{article}
\usepackage{arxiv}
\usepackage[T1]{fontenc}
\usepackage[utf8]{inputenc} 
\usepackage{booktabs}       
\usepackage{amsfonts}       
\usepackage{nicefrac}       
\usepackage{microtype}      
\usepackage{lipsum}
\usepackage{alphabeta}
\usepackage{arydshln}
\usepackage{amsmath,amssymb,amsfonts}
\usepackage{algorithmic}
\usepackage[greek,english]{babel}
\usepackage{cite}
\usepackage{amsmath, amssymb}
\usepackage{multirow}
\usepackage{graphicx}
\usepackage{hyperref}
\usepackage{color}

\title{A Multi-Task Text Classification Pipeline with Natural Language Explanations: A User-Centric Evaluation in Sentiment Analysis and Offensive Language Identification in Greek Tweets\\}

\author{
  Nikolaos Mylonas\\
    Information Technologies Institute,\\ 57001, Greece\\
    \texttt{myloniko@iti.gr}\\
    \And  
  Nikolaos Stylianou\\
    Information Technologies Institute,\\ 57001, Greece\\
    \texttt{nstylia@iti.gr}\\
    \And
  Theodora Tsikrika\\
    Information Technologies Institute,\\ 57001, Greece\\
    \texttt{theodora.tsikrika@iti.gr}\
   \And
   Stefanos Vrochidis\\
    Information Technologies Institute,\\ 57001, Greece\\
    \texttt{stefanos@iti.gr}\\
   \And
   Ioannis Kompatsiaris\\
    Information Technologies Institute,\\ 57001, Greece\\
    \texttt{ikom@iti.gr@iti.gr}\\
}   

\begin{document}
\maketitle      
\begin{abstract}
Interpretability is a topic that has been in the spotlight for the past few years. Most existing interpretability techniques produce interpretations in the form of rules or feature importance. These interpretations, while informative, may be harder to understand for non-expert users and therefore, cannot always be considered as adequate explanations. To that end, explanations in natural language are often preferred, as they are easier to comprehend and also more presentable to end-users. This work introduces an early concept for a novel pipeline that can be used in text classification tasks, offering predictions and explanations in natural language. It comprises of two models: a classifier for labelling the text and an explanation generator which provides the explanation. The proposed pipeline can be adopted by any text classification task, given that ground truth rationales are available to train the explanation generator. Our experiments are centred around the tasks of sentiment analysis and offensive language identification in Greek tweets, using a Greek Large Language Model (LLM) to obtain the necessary explanations that can act as rationales. The experimental evaluation was performed through a user study based on three different metrics and achieved promising results for both datasets.
\end{abstract}

\section{Introduction}
Machine Learning and more specifically Deep Learning models are used in a plethora of different domains and tasks. The predictions provided by such models can often determine the outcome of important processes in many different sectors and may thus impact human livelihood or incur economic costs. In these scenarios, justifications for these predictions are crucial, as they are used by humans to take important decisions, making interpretability essential, in these domains.

Interpretability comes in many different forms, with the most common being rule based and feature importance interpretations~\cite{Saarela2021}. These kinds of interpretations are not always preferred by non-expert users, as they may lack information in case of the former, or not be as intuitive in case of the latter. To that end, explanations in natural language, are becoming increasingly popular, as they are more easily understood by end users, while also containing the necessary information to explain the outcomes of machine/deep learning models. 

This work explores the concept of multi-task pipelines that can provide both predictions and also explanations in natural language. The concept of multi-task predictions, containing both the labels and the corresponding explanations, has been explored mostly for sequence to sequence models~\cite{multi_task1}, in which a single sequence generation model performs both tasks. In our work, we propose a pipeline that can be used to first provide predictions for text classification problems and then combine the predictions with the input to provide explanations in natural language for the predicted label, through the use of a sequence to sequence model. 
Unlike a single multi-task model that generates both labels and explanations simultaneously, our proposed pipeline allows for greater versatility by enabling the use of distinct models for each task, with independent performance, facilitating easier optimisation.

We experiment with our pipeline in a low resource language, namely Greek, for performing sentiment analysis and offensive language detection, two text classification problems, on two datasets originating from X (formerly Twitter) posts~\cite{Tsakalidis2018-ir,offensive}. To obtain the rationales needed to train the sequence to sequence model, we make use of a Greek Large Language Model (LLM) to generate explanations for each instance of the training/validation sets through prompting. Due to the lack of gold standard explanations for both datasets, the performance of our pipeline is evaluated through a user study on the explanations produced by the sequence to sequence model on the test set using three different explainability metrics: \textit{Plausibility}~\cite{plausibility} and \textit{Coherence}~\cite{coherence}, in conjunction with a new metric introduced in this paper referred to as \textit{Perfidiousness}.

Our experimental results show that this pipeline can produce adequate explanations when a sufficient amount of training data with accompanying explanations are available, without hampering the performance of the classifier, as it may happen when a single multi-task model is used. Additionally, the quality of explanations provided by our pipeline does not significantly deteriorate, even when trained on machine generated explanations, rather than ground truth rationales. We make our code and user study results publicly available \footnote{Released upon publication}.

Overall, the main contributions of this work are: (i) a pipeline that can provide both predictions and explanations, effectively performing two different types of downstream tasks, text classification and text generation, with the results of the former being used as input to the later, (ii) a scheme for combining the information of the textual input with the classification label in order to facilitate the learning process of the explanation generator, and (iii) experimental evaluation in a low-resource language where ground truth rationales are not available, and thus we propose a way of obtaining machine-generated explanations that can be used for training the explanation generator.

\section{Related Work}

We first discuss the problems of sentiment analysis and offensive language identification in the Greek language, which are the main focus of this study. A lexicon-based sentiment analysis of Greek tweets during the COVID-19 pandemic made use of a network between accounts to aid in extracting the relevant sentiment of the tweets \cite{su13116150}. Another work focused on sentiment analysis on a scale from 1 to 5 \cite{10.1145/3003733.3003769}, where several Machine Learning models, including a Multi-Layer Perceptron and the RBFKernel, were used for the classification task. To deal with the lack of annotated datasets in the Greek language, a rich set of resources for the Greek language was introduced~\cite{Tsakalidis2018-ir}, ranging from a manually annotated lexicon, to semi-supervised word embedding vectors, and also annotated datasets for different tasks, including a sentiment analysis dataset. This dataset includes the three basic sentiments, namely Positive, Negative, and Neutral, and contains political tweets annotated with these sentiments. Lastly, an aspect-based sentiment analysis on Greek tweets~\cite{stylianou}, using though a private corpus, showed the importance of handling aspect extraction and sentiment analysis end-to-end. Concerning offensive language identification in the Greek language, a recent study~\cite{offensive} introduced a manually annotated dataset containing Greek tweets characterised as containing either offensive or non offensive language. 

Regarding interpretability, most common techniques provide feature importance interpretations, by assigning a weight to each available feature, based on its contribution to the model's output; such techniques include the well-known LIME~\cite{lime}, Integrated Gradients (IG)~\cite{ig}, and SHapley Additive exPlanations (SHAP)~\cite{shap} techniques. Feature importance interpretations, while informative, can sometimes be harder to understand for non expert users, who usually prefer interpretations in natural language. Additionally, such interpretations fail to provide reasoning as to why the model reached a certain prediction, instead simply showcasing the most important parts of the input and their contribution to the decision

To that end, a few works on textual interpretation generation have been recently introduced. Specifically, an approach based on contrastive learning~\cite{meme_detection} makes use of multi-modal inputs of image and text to provide explanations about the meme depicted in the image and whether it constitutes cyberbullying or not.
A framework for generating textual interpretations for visual tasks, called LangXAI~\cite{nguyen2024langxai}, makes use of advanced vision models to provide explanations that are easily understood by human readers and consists of two steps. First, an interpretability technique is used to generate saliency maps for the predictions of a computer vision model; these saliency maps, which are a form of feature importance interpretation, are then given as input, along with the image and ground truth, to a generative model tasked with providing an adequate explanation through prompting.


To evaluate interpretations, human annotated rationales are required. Rationales constitute ground truth explanations, often provided by human experts. When such rationales are available, standard evaluation metrics, such as the F$_1$-Score, can be applied. Additionally, for explanations provided in natural language, the \textit{Simulatability}~\cite{doshivelez2017rigorous} metric can be used, which operates under the notion that a model is considered simulatable, if another model can predict its outputs. In particular, Simulatability evaluates the explanation based on how well a second model can predict the output of the original model, when given the explanation as input. 

However, human annotated rationales are generally hard to find and, therefore, interpretability evaluation can also be performed with unsupervised metrics that usually operate by evaluating certain elements of the interpretation. In the context of feature-importance interpretations, such metrics include \textit{Robustness}~\cite{robustness}, which measures the stability of the interpretation, \textit{Comprehensibility}~\cite{comprehensibility_nzw}, which calculates the percentage of informative features on the interpretation, and \textit{Faithfulness}~\cite{faith}, which measures how faithful the most important feature, according to the interpretation, is to the prediction. Additionally, there is, the \textit{Plausibility}~\cite{plausibility} metric, that is used to evaluate how convincing the interpretation is to humans, regardless of whether the interpretation concerns the correct label or not

When possible, the most appropriate way to evaluate interpretations would be by conducting a user study, as the main purpose of an interpretation is to explain to the user why a certain prediction was made. The standard protocol in these kinds of scenarios is for the users to compare the interpretations provided by various techniques and asses their quality~\cite{user}. However, such human evaluations can be prone to bias, leading to skewed results in certain scenarios~\cite{user2}. 

Models that are able to perform both the task of prediction and the task of explanation are called self-rationalising models~\cite{wiegreffe2022measuring}. Such models are most commonly sequence to sequence classification models trained on both tasks~\cite{multi_task1,multi_task2} and are able to provide plausible free text explanations for different sequence to sequence tasks. Those tasks include natural language inference~\cite{NEURIPS2018_4c7a167b} and machine translation~\cite{ehsan2017rationalization} among others, and for the self-rationalising models to be trained both ground truth labels and ground truth rationales are required. 

The combination of different types of tasks (e.g. classification and free text explanation generation) is less explored. One way of dealing with such scenarios is the employment of pipelines where (i) the model employed first is used for making the prediction for the downstream task, and (ii) the following model is used to generate the rationales based on the input and the prediction. This approach has been used to provide free text explanations for different sequence to sequence tasks including summarization~\cite{wiegreffe2022measuring}, while our work follows this paradigm in order to perform sentiment analysis on Greek tweets, which is a sequence labelling task, and to provide accompanying explanations in natural language.

\section{Text Classification with Natural Language Explanations}

This work introduces a pipeline that produces both predictions and accompanying explanations in natural language for text classification problems. Specifically, we propose a two step process that first classifies a text with a classification model and then combines the predicted label with the input text to generate textual explanations through a sequence to sequence model in a conditional generation approach. This pipeline can be used for any text classification problem, provided that ground truth rationales are available to fine tune the second model.

In particular, the proposed pipeline is comprised of two different models that are trained independently and are used sequentially during inference to provide label predictions along with the explanations. The first model, called the \textit{classifier}, provides the predictions for the text classification task and can be any model capable of handling textual inputs.
The second model, called \textit{explanation generator}, explains in natural language the predictions of the previous model. The input of this model should contain information about the provided input text along with its corresponding label. To that end, we create a new composite text by integrating the information of the input along with its label in the following fashion: ``\{\textit{input text}\} has \{\textit{label}\} label''. This was done to integrate information about the label during the training of the explanation generator, thus resulting in explanations that support the instance's label. Therefore, we produce these composite texts for each instance of the training set and use them to train the \textit{explanation generator}. Additionally, we require ground truth rationales about the instances of the train set in order for the \textit{explanation generator} to be trained.

\begin{figure*}[ht]
    \centerline{\includegraphics[height=2in]{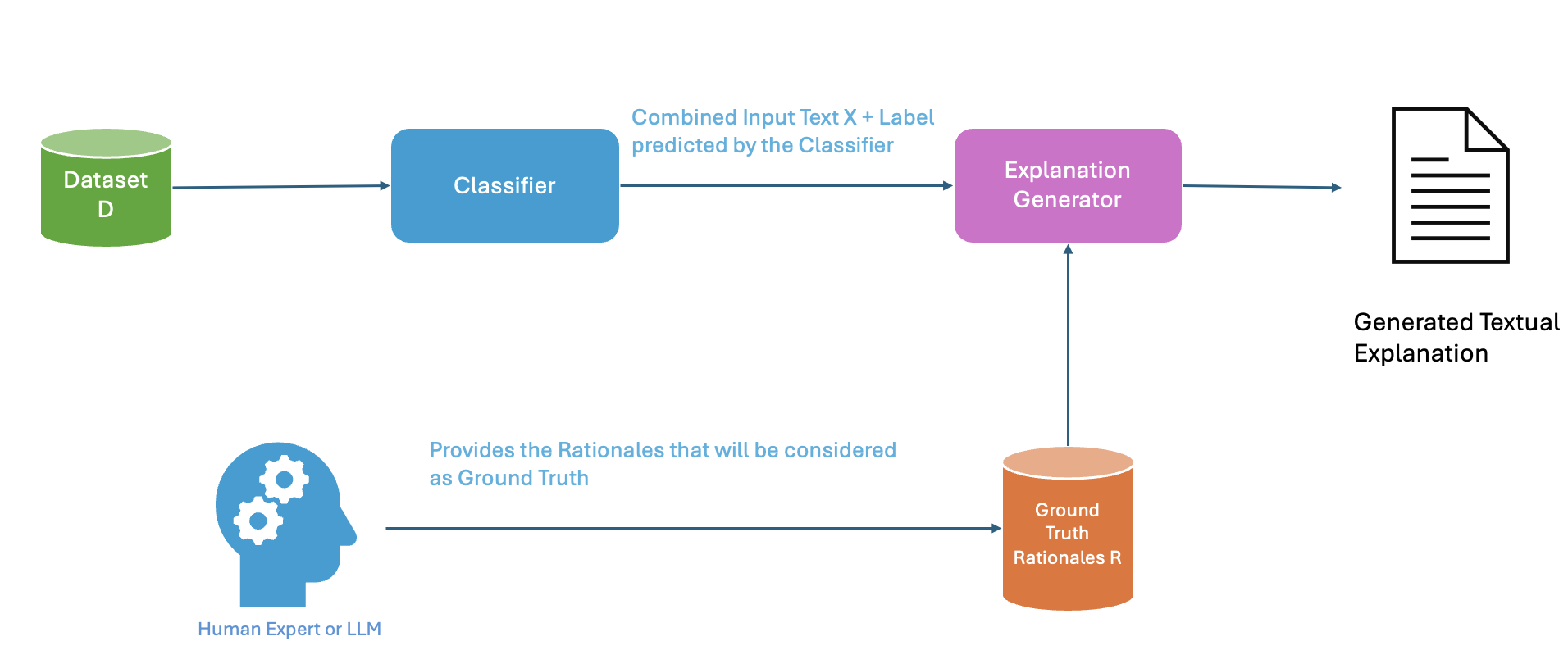}}
    \caption{Proposed pipeline for predictions and natural language explanations}
    \label{fig:pipeline}
\end{figure*}

After the two models comprising the pipeline are trained, the pipeline can be used to predict the label of any incoming text and provide the explanation behind that prediction. Thus, the pipeline not only serves as a tool for classification, but can also be a valuable resource in increasing the transparency of Machine/Deep Learning models. An overview of our our proposed pipeline can be seen in Figure~\ref{fig:pipeline}.

\section{Evaluation Protocol}
This section describes the evaluation protocol followed in this work. Specifically, we describe the datasets used and how we obtain the explanations that act as rationales to build the \textit{explanation generator}, along with our choices for the two models comprising our pipeline. Additionally, we describe the metrics and the process followed to evaluate the quality of the explanations generated by the proposed pipeline through a user-based evaluation.

\subsection{Datasets and Rationale Creation}

While interpretability datasets do exist~\cite{mathew2022hatexplain,eraser}, few provide rationales in natural language and to the best of our knowledge, none of them are for text classification tasks. Therefore, we opt to create rationales for a dataset within a well researched domain, namely that of sentiment analysis, where rationales can add significant value to the predictions.

In particular, we perform our experiments on two datasets one concerning the sentiment analysis of Greek tweets~\cite{Tsakalidis2018-ir}, and the other offensive language identification on Greek tweets~\cite{offensive}. The former dataset contains tweets annotated with one of three sentiments \textit{Positive}, \textit{Negative} or \textit{Neutral}, with the tweets having political content, while the latter contains tweets annotated as either \textit{Offensive} or \textit{Not Offensive}, covering a larger variety of topics. Both datasets were split into training/validation/test sets using the following scheme $70\%/10\%/20\%$. It is worth noting that the distribution of sentiments in the first dataset is heavily imbalanced. Specifically, out of the 1640 total instances in the dataset, only 79 $(4.83\%)$ of them have a positive sentiment, with 582 $(35.48\%)$ having negative, and 979 $(59.69\%)$ having neutral sentiments. Concerning the latter dataset, there is also an inherent imbalance as is common in offensive language identification problem with 2391 $(71.48\%)$ instances out of the 3345 total ones being characterised as containing non-offensive language, and 954 $(28.52\%)$ instances as containing offensive language.

\begin{table}[!ht]
\caption{Original and Translated Prompts used for the Greek LLM Meltemi}
\begin{tabular}{l}
\hline
Conditioning Prompt \\
\hline
\begin{tabular}[c]{@{}l@{}}Θα σου δώσω ένα κείμενο το οποίο έχει \\ χαρακτηριστεί με ένα sentiment. Θέλω να μου επιστρέψει \\
μια πρόταση μόνο που να επεξηγεί τον λόγο \\ για τον οποίο το κείμενο αυτό να χαρακτηριστεί\\ με το sentiment αυτό. Μην γράψεις τίποτα άλλο πέρα\\ από την πρόταση που να επεξηγεί το sentiment.\end{tabular} \\
\hline
Translation\\
\hline
\begin{tabular}[c]{@{}l@{}}I will give you a text that has been labelled with a sentiment. \\ I want you to return only one sentence explaining why \\ this text has been labelled with this sentiment. \\ Do not write anything other than the sentence\\ explaining the sentiment.\end{tabular} \\
\hline
\hdashline
\hline
Query Prompt\\
\hline
\begin{tabular}[c]{@{}l@{}}Το κείμενο: \{input text\} έχει χαρακτηριστεί\\ με το ακόλουθο sentiment \{label\}. \\ Γράψε μου μια πρόταση που να εξηγεί\\ γιατι το κείμενο χαρακτηρίστηκε με το sentiment.\end{tabular} \\
\hline
Translation\\
\hline
\begin{tabular}[c]{@{}l@{}}The text: \{input text\} is labelled \\with the following sentiment \{label\}. \\Write a sentence explaining why the text \\ is labelled with the sentiment.\end{tabular} \\
\hline
\end{tabular}
\label{tab:queries}
\end{table}

To create the rationales needed to train the \textit{explanation generator}, we exploit a generative LLM through a custom created prompts sequence. This approach, even though not ideal, can help us deal with scenarios where ground truth rationales are absent, which is the majority of cases.

In particular, we employ the Greek LLM \textit{Meltemi} (\url{https://www.ilsp.gr/en/news/meltemi-en/}) which is built on top of \textit{Mistral-7B}~\cite{jiang2023mistral} and has been trained on a very large corpus of high-quality Greek texts. We use its instruction-tuned variant \textit{Meltemi-7B-Instruct-v1}, as we want to obtain the explanations through prompting. 

We use two different prompts to obtain the explanations (see Table~\ref{tab:queries}). The first one is used to condition the model so that it returns the output in the desirable format. The second one is the query made to the model, giving it information about the input text and the label, and asks it to return as output the explanation why the label is associated with the input text. Since the model is pre-trained on a Greek corpus, the queries used are constructed using the Greek language; Table~\ref{tab:queries}, however, provides both the original prompt in Greek along with the translated version in English, for ease of understanding. The prompts showcased in Table~\ref{tab:queries}, concern the sentiment analysis dataset, however the ones for the offensive language identification are similar only being slightly different to account for the different type of text classification task.

Using those prompts, we obtain for each instance of the training and validation set the explanations required to train the explanation generator model. Table~\ref{tab:example} provides an example explanation produced by Meltemi for an instance of the sentiment analysis dataset, along with the input text and the label for which the explanation was produced. Yet again, as the input text and the explanation are in Greek, we also provide the translation in English for both; additionally, we remove any mentions of names or political parties, in the showcased samples.

\begin{table}[!ht]
\caption{Input Text, label and Meltemi generated explanation for an instance}
    \begin{tabular}{l}
    \hline
    Original \\
    \hline
    \begin{tabular}[c]{@{}l@{}}
    \textbf{Input Text}: Εκλογές 2015: Επίσημη παρουσίαση του
    \\ \textit{ONOMA} από \textit{KOMMA} και \textit{ΚΟΜΜΑ}\\  
    \textbf{Sentiment}: Neutral\\
    \textbf{Explanation}: Το κείμενο παρουσιάζει τον \textit{ONOMA}\\
    ως υποψήφιο για τις εκλογές του 2015,\\
    χωρίς να εκφράζει καμία θετική ή αρνητική γνώμη για αυτόν.\\ Επομένως, χαρακτηρίζεται ως ουδέτερο.\end{tabular} \\
    \hline
    Translation\\
    \hline
    \begin{tabular}[c]{@{}l@{}}
    \textbf{Input Text}: 2015 Elections: official presentation of \textit{NAME}\\
    by \textit{POLITICAL PARTY} and the \textit{POLITICAL PARTY}\\
    \textbf{Sentiment}: Neutral\\
    \textbf{Explanation}: The text presents \textit{NAME} as \\
    a candidate for the 2015 elections,\\ without expressing any positive or negative opinion about them. \\ It is therefore described as neutral. \end{tabular} \\
    \hline
    \end{tabular}
    \label{tab:example}
\end{table}

\subsection{Model Selection}

As our experimental evaluation focuses on two different text classification tasks in Greek tweets, we decided to use Greek-BERT~\cite{greek_bert}  for the first model (\textit{classifier}). Greek-BERT is a BERT~\cite{devlin2019bert} model, pre-trained on a large corpus of Greek data, including texts from the Greek part of Wikipedia and European Parliament records in Greek. Once the model is fine-tuned on a task-specific dataset, it can then be used to provide predictions for incoming text instances. It should be noted that the \textit{classifier} was fine-tuned for 15 epochs for the sentiment analysis dataset, and for 10 epochs for the offensive language identification dataset.

For the second model (\textit{explanation generator}), we chose BART~\cite{lewis2019bart}, a model that has been widely used for different sequence to sequence tasks, including summarisation and machine translation. BART is easy to implement and fine-tune, in addition to its adequate performance in different datasets, thus making it a solid option. Our \textit{explanation generator} was fine-tuned for 15 epochs for both datasets.

\subsection{User-Centred Evaluation}

As the proposed pipeline consists of two models, each accomplishing a different task, we evaluate each model separately. Additionally, since the tasks of sentiment analysis and offensive language identification have gold standard annotations in the dataset, our user-centred study is focused only on evaluating the second part of the pipeline, namely the generated explanations, for which no ground truth rationales are available. To that end, we used the F1-Score and Balanced Accuracy metrics to evaluate the performance of the classifier part of the pipeline, while for the explanation generator, we conducted a user-centred evaluation through a user study.

As there are no ground truth rationales for the datasets we used, we cannot evaluate our explanations using metrics such as Simulatability~\cite{doshivelez2017rigorous}. Additionally, since we have explanations in natural language, we cannot use unsupervised metrics such as faithfulness-based metrics. Based on these facts, the most appropriate way to evaluate our explanations is through a user study.

We evaluate the quality of the created explanations through a user study using three different metrics. Specifically, we employ the \textit{Plausibility} metric, designed for natural language explanations. This metric quantifies how plausible the provided explanation is to the predicted sentiment according to the input text. A plausible explanation should explain to the user the reason why the sentiment was selected for that input text, even if that sentiment is incorrect. Based on that, this metric tries to quantify how convincing the explanation is to the end user.

The second metric is called \textit{Coherence} and quantifies how coherent the explanation is. A coherent explanation should be very close to a human written text, being devoid of syntactical or grammatical errors. On the other hand, an explanation with very low coherence would be akin to a text with randomly placed words, without any context or meaning.


For the final metric, we propose \textit{Perfidisouness} which aims to quantify how good a generated explanation is for each instance, concerning a label other than the predicted one. Specifically, an explanation with high Perfidiousness should be faithful to any label other the predicted one, while achieving low values when the explanation is faithful to the predicted label. As a result, Perfidisouness measures the \textit{explanation generator}`s ability to capture label information from the textual context, ignoring the provided predicted label. 

For example, let us suppose that we have a textual input that was predicted by our pipeline to have a Neutral sentiment. In case the explanation accompanying that prediction describes the example as Neutral and the reasons why it was predicted as such, Perfidiousness would be low. On the other hand, in case the explanation interprets the instance as Positive or Negative, giving also a reasoning as to why that is, Perfidiousness would be high.
We use Perfidiousness to gauge the effectiveness of the proposed label conditional prompting approach and simultaneously as an indicator for the necessity of the classification task in the pipeline; in particular, if our explanations exhibit high Perfidisouness, there would be no need to integrate label information during the explanation generation process.

Two user studies were performed one for each dataset, with the same number of participants, specifically 15 native speakers all of whom had a technical background with some knowledge of machine/deep learning, but not necessarily explainability. We planned for the user study to be performed on 10 random instances per label on the corresponding dataset, resulting in 30 instances for the sentiment analysis one and 20 for the offensive language identification. However, the test set for the sentiment analysis dataset, contained only 8 examples for the positive sentiment. Based on that, to keep a balanced distribution of the selected instances, we randomly selected 11 instances with neutral and 11 instances with negative as the predicted sentiments, along with the 8 ones that had the positive sentiment. Since the number of instances per label is larger in the offensive language identification dataset, we did not face any such issue in that case. 

The same instances were presented to all users for both user studies, giving them details about the input text, the label, and the explanation. The users rated the generated explanation according to each metric, using a score from 1 to 10 (with higher values denoting better performance). The scores of the metrics were then computed as the average score each user gave for each instance and then the average across all the instances was computed.

\section{Experimental Results}
\begin{table}[!t]
\caption{Average performance of explainability metrics per examined sentiment}
\centering
    \begin{tabular}{cccc}
    \hline
    Sentiment & Plausibility & Coherence & Perfidiousness\\
    \hline
    Neutral & 9.00 & 7.68& 1.41\\
    Negative & 6.34 & 5.06 & 2.41\\
    Positive & 5.46 & 4.29 & 2.96 \\
    \hline
    Overall & 7.08 & 5.82 & 2.19 \\
    \hline
    \end{tabular}
    \label{tab:results_per}
\end{table}

It is worth noting, that our experimental results do not focus as much in the classification task, as both text classification problems studied in this work are generally considered as 'solved' with the right tuning. Instead, this work mostly focuses on the evaluation of the generated textual explanations. Having said that, we will first showcase the results for the sentiment analysis dataset, based on the Balanced Accuracy and macro averaged F1-Score metrics. We additionally, showcase the F1-Score for each specific sentiment. The performance is the following: Balanced Accuracy $(0.929)$, F1 $(0.798)$, $F1_{Pos}$ $(0.583)$, $F1_{Neg}$ $(0.875)$, $F1_{Neu}$ $(0.936)$. We can see that, as expected, the model performs much better for the Neutral (Neu) and Negative (Neg) sentiments, as there are a lot more available training data for these two sentiments. On the contrary, the performance shows a significant drop for the Positive (Pos) sentiment, where only a small amount of training data are available.

Regarding the evaluation of the explanations the average value for each metric according to the user study is presented at Table~\ref{tab:results_per}, for each different sentiment and overall.We can see that the provided explanations seem to have pretty high Plausibility and relatively average Coherence, according to the different users. Concerning the Perfidiousness metric, the value reported by the users seems to be pretty low, leading us to believe that in most cases our explanations are faithful to the predicted sentiment and not on different ones. 

Due to the imbalance of the dataset for certain sentiments, there are not enough training data to train the explanation generator adequately, and therefore it is expected that the explanations provided for those sentiments are of inferior quality. The provided explanations concerning the Neutral sentiment, which is the one with the most examples in the training set, seem to be of higher quality for both Plausibility and Coherence, when compared to the other two sentiments. This is expected as the sequence to sequence model has encountered more examples with that sentiment during training, and therefore was able to better understand how to provide explanations for it. Similarly, the performance for the Negative Sentiment seems to be better when compared to the Positive sentiment, due to the very low number of instances labelled with the latter in the training set.

For the offensive language identification dataset, the results of the classifier are the following: Balanced Accuracy $(0.866)$, F1 $0.855$, $F1_{Not Off}$ $(0.91)$, $F1_{Off}$ $(0.8)$. Similarly to before, the performance of the classifier is high even without any sophisticated tuning procedure. This time however, unlike the previous dataset the model performs similarly for both labels, as there is not a noticeable imbalance in the dataset. For the evaluation of the interpretations, the average value for each metric can be found in Table~\ref{tab:results_per_off}. We can see that the explanations provided for the language identification dataset, seem to have higher overall values for Plausibility and Coherence metrics. This is expected, as both those labels had an adequate amount of train examples for the \textit{explanation generator}, to learn how to provide suitable explanations. Additionally, the Perfidiousness metric has a lower value for this dataset according to the users, meaning that more training data result in explanations that are more faithful to the predicted label, as the generator is able to better distinguish between the different labels.

\begin{table}[!t]
\caption{Average performance of explainability metrics for offensive and not offensive labels}
\centering
    \begin{tabular}{cccc}
    \hline
    Sentiment & Plausibility & Coherence & Perfidiousness\\
    \hline
    Offensive & 7.93 & 7.50 & 1.83 \\
    Not Offensive & 8.45 & 7.37 & 1.67\\
    \hline
    Overall & 8.19 & 7.44 & 1.75 \\ 
    \hline
    \end{tabular}
    \label{tab:results_per_off}
\end{table}

All in all, we can see that the explanations provided by our pipeline seem to have high Plausibility, especially when concerning a label with an adequate amound of training data, meaning that they are able to capture and explain the predicted label of each text. Additionally, they have relatively high Coherence, meaning they provide a text that is understood by humans, but may not always be as clear as a human written text. The Coherence of the explanations seems to yet again increase as a higher amount of training data is provided to the generative model. Finally, the Perfidiousness of the generated explanations is low in both datasets, meaning that the provided explanations do not represent any other label besides the predicted one. An exception appears in the case of the sentiment analysis dataset, and specifically for the Positive sentiment, in which a low relative F1 score is achieved by the \textit{classifier}

\section{Conclusions}

In this paper, we presented a pipeline that can provide both predictions and the corresponding explanations for text classification problems. This pipeline consists of two separate models, one classifier for providing the predictions and one sequence to sequence model which produces the explanations, and can be used to provide transparency in domains where text classification is used.

Through experimentation, we found that the explanations provided by the pipeline contain sufficient information for users, while being coherent for the most part, meaning that they resemble human written text. Additionally, as the classifier model is trained independently of the explanation generator, the performance of the model in the downstream task of text classification does not deteriorate. Additionally, it was found that the quality of explanations increases, when more training data are available. This is expected, as the \textit{explanation generator}, has more available data and rationales to learn from, and therefore is able to produce not only more plausible explanations, but more coherent ones as well.

As future work, we aim to extend our experimental procedure to include a larger number of datasets from the English language, ideally ones where human annotated rationales are present. Based on our initial findings, we hope that with more available data, the quality of the produced explanations will increase both in Plausibility, as well as Coherence, as was indicated by the higher quality explanations produced for the Neutral sentiment and the offensive language identification dataset, where more data were available.

A larger user study that includes more human users from different backgrounds will also be conducted to have a more representative sample of real users for our evaluation. Furthermore, we will also experiment with using a single self-rationalising model that can provide the predictions and explanations at the same time to compare how the quality of predictions and explanations changes. Finally, testing different models for the classifier and explanation generator is another future research direction we plan to tackle.

\bibliographystyle{unsrt}

\end{document}